\documentclass[letterpaper]{article} 
\usepackage[preprint]{aaai2027}  
\usepackage[hyphens]{url}  
\usepackage{graphicx} 
\usepackage{natbib}  
\usepackage{caption} 
\usepackage{algorithm}
\usepackage{algorithmic}

\usepackage{newfloat}
\usepackage{listings}
\DeclareCaptionStyle{ruled}{labelfont=normalfont,labelsep=colon,strut=off} 
\floatstyle{ruled}
\newfloat{listing}{tb}{lst}{}
\floatname{listing}{Listing}

\usepackage{booktabs}
\usepackage{graphicx} 
\usepackage{textcomp}
\usepackage{amssymb}
\title{pseudocode}

\usepackage{times}
\usepackage{latexsym}
\usepackage{graphicx}
\usepackage{makecell}
\usepackage{colortbl}
\usepackage{pifont}

\usepackage[T1]{fontenc}
\usepackage{xcolor}
\usepackage[utf8]{inputenc}

\usepackage{microtype}

\usepackage{pifont}
\usepackage{tabularx}
\usepackage{adjustbox}
\usepackage{multirow}
\usepackage{enumitem}
\usepackage{xspace}
\usepackage{tcolorbox}
\usepackage{xparse}
\usepackage{soul}
\usepackage{booktabs,dcolumn}
\usepackage{url}
\usepackage{amsthm,amssymb,bm,stmaryrd}
\usepackage{breqn}
\usepackage{arydshln}

\usepackage{arydshln} 
\usepackage{makecell}
\usepackage{booktabs}
\usepackage{amsmath}
\usepackage{color}
\usepackage[table]{xcolor}
\definecolor{yellow_l}{RGB}{191,144,0}
\definecolor{blue_l}{RGB}{0,176,240}
\definecolor{blue_case}{RGB}{16,111,183}
\definecolor{yellow_case}{RGB}{255,230,153}
\definecolor{colored_case1}{RGB}{0,86,255}
\definecolor{colored_case2}{RGB}{255,0,0}
\definecolor{colored_case3}{RGB}{154,243,0}
\definecolor{colored_case4}{RGB}{248,115,23}
\definecolor{colored_case5}{RGB}{178,74,255}
\definecolor{colored_case6}{RGB}{255,154,212}
\definecolor{colored_case7}{RGB}{127,96,0}
\usepackage{multirow}
\usepackage{enumitem}
\usepackage{bbding}
\definecolor{lightblue}{RGB}{232,242,255}

\definecolor{a}{RGB}{251, 229, 214}
\definecolor{b}{RGB}{222, 235, 247}
\definecolor{c}{RGB}{226, 240, 217}

\definecolor{hui}{RGB}{242, 242, 242}
\definecolor{lv}{RGB}{210, 255, 218}

\definecolor{lan}{RGB}{30, 119, 179}
\definecolor{zi}{RGB}{112, 48, 159}
\definecolor{cheng}{RGB}{255, 127, 15}

\usepackage{tcolorbox}
\usepackage{xcolor}
\tcbuselibrary{breakable, skins}

\tcbset{
  promptbox/.style={
    breakable,
    enhanced,
    colback=gray!5!white,
    colframe=gray!40!black,
    colbacktitle=gray!20!white,
    coltitle=black,
    fonttitle=\small\bfseries,
    title={#1},
    boxrule=0.5pt,
    arc=4pt,
    left=6pt, right=6pt, top=4pt, bottom=4pt,
    attach boxed title to top left={yshift=-2mm, xshift=6pt},
    boxed title style={
      colback=gray!20!white,
      colframe=gray!40!black,
      boxrule=0.5pt,
      arc=3pt,
    },
  }
}

\newtcolorbox{takeaways}{
  colback=blue!5,
  colframe=blue!45!black,
  boxrule=0.5pt,
  arc=2pt,
  left=6pt, right=6pt, top=4pt, bottom=4pt
}
\title{Uncovering Shortcut Amplification in Video RLVR through the \\ Lens of Perceptual Uncertainty Collapse}
\author {
    Jiafeng Liang\textsuperscript{\rm 1}\equalcontrib,
    Xianhao Ou\textsuperscript{\rm 1}\equalcontrib,
    Tao Ren\textsuperscript{\rm 4}\equalcontrib,
    Rongchuan Mu\textsuperscript{\rm 1},
    Runxuan Liu\textsuperscript{\rm 1},
    Dandan Tu\textsuperscript{\rm 5},\\
    Ming Liu\textsuperscript{\rm 1, \rm 2}\corresponding,
    See-Kiong Ng\textsuperscript{\rm 3},
    Bing Qin\textsuperscript{\rm 1, \rm 2}
}
\affiliations {
    \textsuperscript{\rm 1}Harbin Institute of Technology,
    \textsuperscript{\rm 2}PengCheng Laboratory,
    \textsuperscript{\rm 3}National University of Singapore,\\
    \textsuperscript{\rm 4}Peking University,
    \textsuperscript{\rm 5}Huawei Technologies Co., Ltd\\
    jfliang@ir.hit.edu.cn, xhou@ir.hit.edu.cn, mliu@ir.hit.edu.cn
}

\begin{document}

\maketitle

\begin{abstract}

Reinforcement learning with verifiable rewards (RLVR) has emerged as an effective post-training paradigm for enhancing the video reasoning capability of Large Multimodal Models (LMMs). 
Despite its empirical effectiveness, we show that outcome-level optimization can introduce an undesirable shortcut amplification effect, whereby models increasingly favor text-prior answers even when they are not supported by the visual evidence. 
To characterize this behavior, we investigate the training dynamics of video RLVR through the lens of perceptual uncertainty, analyzing how predictive uncertainty evolves under video-conditioned and text-only contexts. 
Our analysis identifies a \textbf{perceptual uncertainty collapse} phenomenon, in which RL training continuously sharpens the model's answer distribution while the uncertainty reduction attributable to visual evidence progressively diminishes. 
A further gradient-direction analysis reveals that such confidence enhancement becomes increasingly less aligned with visually grounded uncertainty reduction and is instead dominated by text-prior compression. 
Motivated by these findings, we propose \textbf{B}ifocal \textbf{S}elf-distillation \textbf{P}olicy \textbf{O}ptimization (\textbf{BSPO}), a self-distillation framework that regulates policy updates with two complementary distributional signals. 
Specifically, BSPO derives empirical reasoning principles from intra-group correct and incorrect rollouts to identify reasoning-sensitive tokens, and introduces shortcut-aware visual perturbation to penalize predictions that remain invariant when task-relevant visual evidence is weakened. 
By adaptively reweighting the original RL advantage, BSPO preserves the task-solving benefits of RLVR while promoting updates that are both reasoning-sensitive and visually grounded. 
Extensive experiments validate the effectiveness of BSPO in improving video reasoning performance and strengthening visual faithfulness.

\end{abstract}

\section{Introduction}

\begin{figure}[t]
    \centering
    \includegraphics[width=\linewidth]{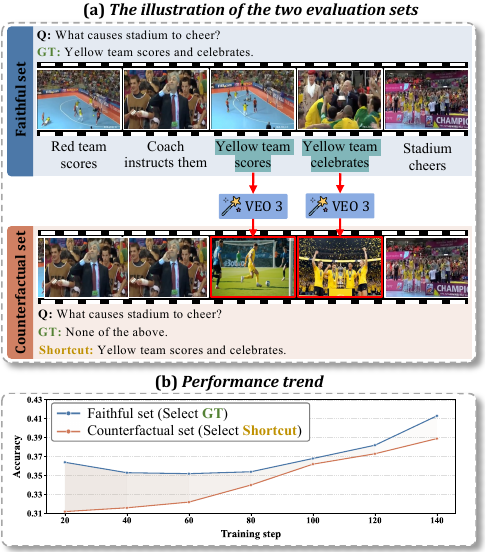}
    \caption{Training dynamics observations of video RLVR. (a) The illustration of the two evaluation sets. (b) Accuracy of selecting the ground-truth (GT) answer on the faithful set and the shortcut answer on the counterfactual set.}
    \label{fig:motivation}
\end{figure}

Reinforcement learning with verifiable rewards (RLVR) has become a widely used post-training paradigm~\cite{deepseekr1,dapo} for improving the reasoning ability of Large Multimodal Models (LMMs)~\cite{mimovl,qwen3vl,kimivl}. 
By using rule-based outcome rewards, a model is encouraged to generate responses that match the final answer, while the optimization process does not require dense human annotations for intermediate reasoning steps. 
In this setting, RLVR enables LMMs to learn from automatically verifiable supervision and has shown promising potential in video reasoning tasks. 

Unlike text-only reasoning, video reasoning requires generating reliable predictions grounded in temporal visual information.
Recent studies~\cite{zhihao,perception} have shown that LMMs may ignore visual evidence during reasoning and instead rely on shortcuts derived from prior knowledge or textual contextual cues.
However, we observe that this phenomenon is not entirely inherent to the model, but can be induced by RL training.
As shown in Figure~\ref{fig:motivation}(a), we train a baseline LMM~\cite{qwen3vl} with standard GRPO and evaluate checkpoints collected throughout training.
At each checkpoint, we evaluate the model on two test sets, one of which uses the original videos as the faithful set, while the other replaces key video segments with semantically related but visually irrelevant clips generated by a video generator~\cite{veo3}.
In the counterfactual set, the question and options remain unchanged, but the original ground-truth answer becomes a shortcut answer favored by textual priors, while “None of the above” is added as the new correct answer. 
Figure~\ref{fig:motivation}(b) illustrates a gradually narrowing performance gap between the two sets during training, indicating that shortcut predictions are progressively reinforced even when they are not supported by the visual evidence.

To systematically investigate this observation, we analyze the training dynamics of video RLVR from a perceptual uncertainty perspective. 
Specifically, we compute predictive entropy under both video-conditioned and text-only inputs, and introduce the visual-conditioning entropy gap to quantify the extent to which uncertainty reduction can be attributed to visual evidence. 
If RL optimization genuinely enhances visual utilization, the reduction in video-conditioned entropy should be accompanied by an increasing entropy gap, indicating that the model becomes more certain by extracting information from the video. 
However, our results reveal a contradictory pattern, which we term perceptual uncertainty collapse, whereby RL optimization progressively compresses multiple plausible visual interpretations into an overconfident prediction that is insufficiently grounded in the video.
To further characterize its underlying mechanism, we conduct a gradient-direction analysis of RL-induced uncertainty compression. 
Our analysis shows that outcome-level RL updates can sharpen next-token distributions by aligning the policy update direction with entropy-reducing directions. 
However, this sharpening effect is not necessarily visually specific. 
As training proceeds, visual-specific compression gradually weakens, indicating that RL-induced confidence becomes increasingly decoupled from visual evidence and is instead dominated by text-prior compression.

Based on the above analysis, we propose Bifocal Self-distillation Policy Optimization (BSPO), a self-distillation framework that aims to retain the reasoning benefits of RLVR while suppressing shortcut-driven optimization. 
BSPO consists of two complementary components. 
First, Principle-guided Credit Assignment (PCA) derives empirical reasoning principles by contrasting correct and incorrect trajectories within the same rollout group. 
These principles serve as guidance for identifying reasoning-sensitive tokens, thereby enabling more fine-grained credit assignment than standard trajectory-level advantage estimation. 
Second, Shortcut-aware Contrastive Perturbation (SCP) weakens task-relevant visual evidence and compares the resulting policy distribution with that under the original video input. 
SCP penalizes predictions that remain overly invariant to visual perturbation, since such invariance indicates weak visual dependence.
Finally, BSPO incorporates these two signals into a bifocal optimization objective that adaptively modulates the magnitude of the original RL advantage without altering its direction. 
In this way, our method preserves the task-solving gains of RLVR while encouraging updates that are both reasoning-sensitive and visually grounded.
Extensive experiments demonstrate that our BSPO outperforms mainstream standard RLVR baselines. 
Further analysis shows that BSPO preserves the visual-conditioning entropy gap and mitigates visual-specific compression decline, indicating more visually faithful reasoning.

\section{Rethinking Video RLVR from a Perceptual Uncertainty Perspective}

\subsection{Background and Motivation}

We formalize the video RLVR post-training problem as follows. 
Consider a video-question pair $(v, q)$ drawn from dataset $\mathcal{D}$, where $v$ represents the visual modality and $q$ denotes the corresponding textual query. 
The model $\pi_\theta$ with parameters $\theta$ generates a response $o \sim \pi_\theta\left (\cdot \mid v, q\right )$ and each response is assessed by a rule-based verifier $R(\cdot)$, which provides a binary reward of $1$ for a correct answer and $0$ for an incorrect one. 
The training objective is to maximize the expected reward:
\begin{equation*}
    \mathcal{J}(\theta) = \mathbb{E}_{(v,q) \sim \mathcal{D},\ o \sim \pi_\theta(\cdot \mid v, q)}\left[R(o)\right]
\end{equation*}
For widely adopted baseline GRPO~\cite{deepseekr1}, given a group of $G$ responses $\{o_i\}_{i=1}^G$ sampled from a reference model $\pi_{\theta^{'}}\left (\cdot \mid v, q\right )$, the objective is defined as:
\begin{equation*}
\begin{split}
\mathcal{J}_{\text{GRPO}}(\theta) = \mathbb{E}_{\substack{(v,q)\sim\mathcal{D}\\\{o_i\}_{i=1}^G \sim \pi_{\theta'}(\cdot|v,q)}}
\Bigg[ \frac{1}{G}\sum_{i=1}^{G} \frac{1}{|o_i|}\sum_{t=1}^{|o_i|} \\
\min\!\left(w_{i,t}(\theta)A_i,\ \text{clip}\!\left(w_{i,t}(\theta), 1-\epsilon, 1+\epsilon\right)A_i\right)\Bigg],
\end{split}
\end{equation*}
where the group-normalized advantage is $A_i = \tfrac{R(o_i) - \mathrm{Mean}(R(o))}{\text{Std}(R(o))}$ and $w_{i,t}(\theta) = \tfrac{\pi_\theta(o_{i,t} \mid v, q, o_{i,<t})}{\pi_{\theta'}(o_{i,t} \mid v, q, o_{i,<t})}$ is the token-level importance sampling ratio.
The scalar reward $R(o) \in \{0, 1\}$ only measures whether the model obtained the correct answer and output format.

Figure~\ref{fig:motivation} reveals a counter-intuitive trend in which the performance gap between video-conditioned and text-only inputs gradually narrows as RL training progresses.
This raises a natural question: \emph{Do the performance gains reflect stronger visual grounding, or merely increased confidence driven by textual priors?}
To answer this question, we investigate how predictive uncertainty evolves throughout RL training.

\begin{figure*}[t]
    \centering
    \includegraphics[width= \linewidth]{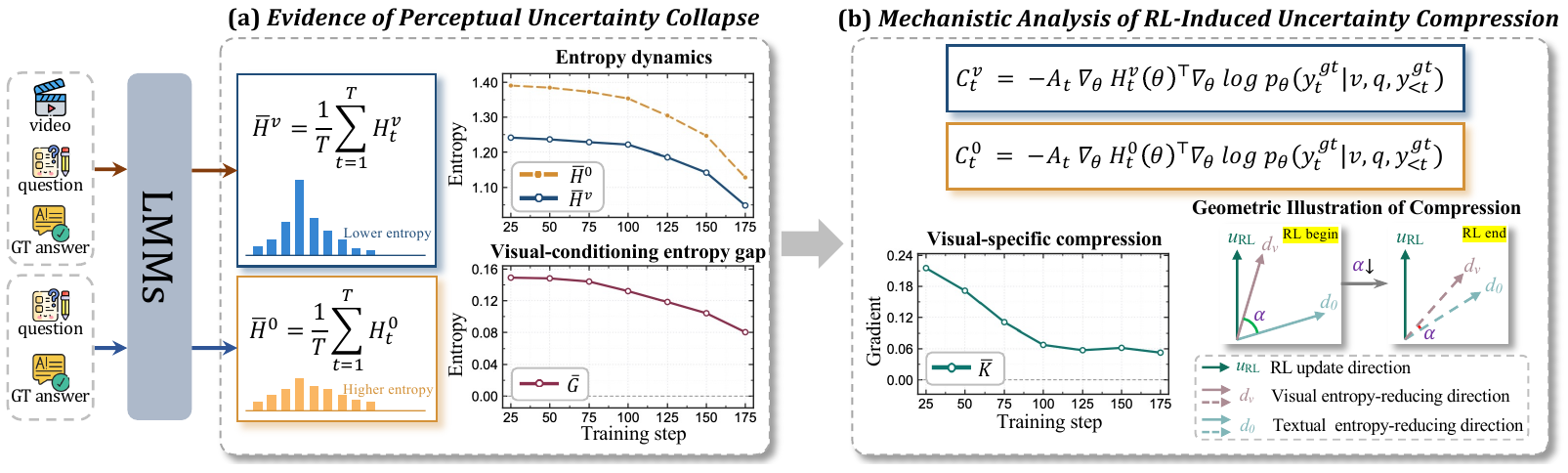}
    \caption{Perceptual uncertainty analysis of video RLVR. (a) RL training reduces the video-conditioned entropy $\bar{H}^v$, while the visual-conditioning entropy gap $\bar{G}$ also decreases, indicating that increased confidence is not accompanied by stronger visual evidence utilization. 
    (b) The decline of visual-specific compression $\bar{K}$ and geometric misalignment, suggests that such sharpening is progressively decoupled from visual entropy-reducing directions and increasingly dominated by text-prior compression.}
    \label{fig:analysis}
\end{figure*}

\subsection{Evidence of Perceptual Uncertainty Collapse}

Given a video-question pair $(v,q)$ and its ground-truth response $y^{\mathrm{gt}}=(y^{\mathrm{gt}}_1,\ldots,y^{\mathrm{gt}}_T)$, we compute the next-token predictive distribution under teacher forcing.
For the $t$-th position, the video-conditioned entropy is defined as
\begin{equation*}
H_t^{v}
=
-\sum_{w\in\mathcal{V}}
p_\theta\!\left(w\mid v,q,y_{<t}^{\mathrm{gt}}\right)
\log p_\theta\!\left(w\mid v,q,y_{<t}^{\mathrm{gt}}\right),
\end{equation*}
where $\mathcal{V}$ denotes the vocabulary. 
We use the ground-truth response under teacher forcing rather than model-generated rollouts to ensure a controlled comparison between video-conditioned and text-only contexts. 
Similarly, we compute the text-only entropy by removing the video input:
\begin{equation*}
H_t^{0}
=
-\sum_{w\in\mathcal{V}}
p_\theta\!\left(w\mid q,y_{<t}^{\mathrm{gt}}\right)
\log p_\theta\!\left(w\mid q,y_{<t}^{\mathrm{gt}}\right).
\end{equation*}
We then average the token-level entropy over the trajectory:
\begin{equation*}
\bar{H}^{v}
=
\frac{1}{T}\sum_{t=1}^{T} H_t^{v},
\qquad
\bar{H}^{0}
=
\frac{1}{T}\sum_{t=1}^{T} H_t^{0}.
\end{equation*}
To measure the uncertainty reduction attributable to visual evidence, we define the visual-conditioning entropy gap as:
\begin{equation*}
G_t
=
H_t^{0}-H_t^{v},
\qquad
\bar{G}
=
\frac{1}{T}\sum_{t=1}^{T}G_t.
\end{equation*}
Here, $\bar{H}^{v}$ reflects the model's predictive uncertainty under visual conditioning, while $\bar{G}$ quantifies the reduction in predictive uncertainty induced by visual input. 
If RL optimization improves visual utilization, a decrease in $\bar{H}^{v}$ should be accompanied by an increase in $\bar{G}$.

However, as shown in Figure~\ref{fig:analysis}(a), $\bar{H}^{v}$ consistently decreases as training progresses, whereas $\bar{G}$ exhibits no improvement and instead declines consistently.
This indicates that the model becomes increasingly confident without extracting proportionally more information from the video. We refer to this mismatch as \emph{perceptual uncertainty collapse}.

\begin{takeaways}
	\textbf{Takeaway: } 
	The decrease in video-conditioned entropy does not indicate improved visual utilization.
\end{takeaways}\vspace{-5pt}

\subsection{Mechanistic Analysis of RL-Induced Uncertainty Compression}
To further characterize the mechanism underlying this failure mode, we examine it from a gradient-geometric perspective and show how outcome-level RL can naturally induce uncertainty collapse.
For the $t$-th position in the ground-truth trajectory, we first consider the video-conditioned predictive entropy $H_t^v$.
Ignoring clipping terms for simplicity, the token-level GRPO objective can be formalized as:
\begin{equation*}
J_t(\theta)
=
A_t
\log p_\theta(y_t^{\mathrm{gt}} \mid v,q,y_{<t}^{\mathrm{gt}}),
\end{equation*}
where $A_t$ is the advantage assigned to this token.
The corresponding first-order update direction is:
\begin{equation*}
\theta^{+}
=
\theta
+
\eta A_t
\nabla_\theta
\log p_\theta(y_t^{\mathrm{gt}} \mid v,q,y_{<t}^{\mathrm{gt}}),
\end{equation*}
where $\eta$ is the learning rate.
We then approximate the entropy change after one update step:
\begin{align*}
H_t^v(\theta^{+}) - H_t^v(\theta)
&\approx
\nabla_\theta H_t^v(\theta)^\top(\theta^{+}-\theta) \nonumber \\
&=
\eta A_t
\nabla_\theta H_t^v(\theta)^\top
\nabla_\theta
\log p_\theta(y_t^{\mathrm{gt}} \mid v,q,y_{<t}^{\mathrm{gt}}).
\end{align*}
This expression shows that the entropy change is governed by the geometric alignment in parameter space between the entropy gradient and the policy-gradient update direction.
To make positive values correspond to entropy reduction, we define the uncertainty compression score:
\begin{equation*}
C_t^v
=
-
A_t
\nabla_\theta H_t^v(\theta)^\top
\nabla_\theta
\log p_\theta(y_t^{\mathrm{gt}} \mid v,q,y_{<t}^{\mathrm{gt}}).
\end{equation*}
A positive value of $C_t^v$ indicates that the RL update reduces the video-conditioned entropy $H_t^v$, thereby sharpening the next-token distribution under visual input. 
Geometrically, $C_t^v$ measures the relative projection strength of the RL update onto the video-conditioned and text-only entropy-reducing directions.
However, such entropy compression does not necessarily imply stronger visual grounding, since the same update may also align with the text-only entropy-reducing direction. 
Therefore, we further compute the text-only counterpart $C_t^0$:
\begin{equation*}
C_t^0
=
-
A_t
\nabla_\theta H_t^0(\theta)^\top
\nabla_\theta
\log p_\theta(y_t^{\mathrm{gt}} \mid v,q,y_{<t}^{\mathrm{gt}}),
\end{equation*}
and compared with $C_t^v$ to quantify the relative compression effect, termed visual-specific compression:
\begin{equation*}
K_{t} = C_t^v - C_t^0,
\qquad
\bar{K} = \frac{1}{T}\sum_{t=1}^{T}K_{t} .
\end{equation*}
If $C_t^v$ is substantially larger than $C_t^0$, the update is more consistent with visually grounded confidence formation. 
In contrast, the sharpening effect is more likely driven by generic text-prior reinforcement rather than additional visual evidence.

As shown in Figure~\ref{fig:analysis}(b), although RL updates induce uncertainty compression and thus push the model toward sharper predictive distributions, the visual-specific compression gradually diminishes. 
Together with the declining visual-conditioning entropy gap, this suggests that the increased confidence becomes less attributable to visual input and increasingly reflects generic text-prior sharpening, providing a mechanistic explanation for perceptual uncertainty collapse.
\begin{takeaways}
	\textbf{Takeaway: } 
	Outcome-level RL can progressively amplify overconfidence by sharpening text-prior predictions, without necessarily improving visual perception.
\end{takeaways}\vspace{-5pt}

\section{Bifocal Self-distillation Policy Optimization}
To mitigate visual uncertainty collapse in video RLVR, we propose Bifocal Self-distillation Policy Optimization (BSPO), which integrates Principle-guided Credit Assignment (PCA) and Shortcut-aware Contrastive Perturbation (SCP) to improve reasoning capability while preserving dependence on visual evidence.
Instead of relying on external annotations or an independent teacher model, the proposed objective derives supervision from the model's own distributional variations under different input conditions. 
Specifically, PCA contrasts correct and incorrect trajectories within the same rollout group to derive retrospective principles that guide subsequent reasoning refinement (Section~~\ref{ssec:pca}).
Meanwhile, SCP compares the model's predictions under original and perturbed video inputs, penalizing responses that remain overly stable when visual evidence is weakened (Section~\ref{ssec:scp}). 

\subsection{Principle-guided Credit Assignment}
\label{ssec:pca}

In standard video RLVR, the same trajectory-level advantage is assigned to all tokens, causing decisive reasoning steps and generic language tokens to receive identical credit.
Principle-guided Credit Assignment (PCA) addresses this limitation by using empirical principles as probes to identify reasoning-sensitive tokens, enabling more precise credit assignment and improved reasoning performance.

Given a video-question pair $(v,q)$, the policy samples a group of responses $\mathcal{O}=\{o_i\}_{i=1}^{G}$. 
According to the reward $R(o_i)$, we split the group into correct and incorrect subsets:
\begin{equation*}
    \mathcal{O}^{+}=\{o_i\in\mathcal{O}:R(o_i)>0\},
    \;
    \mathcal{O}^{-}=\{o_i\in\mathcal{O}:R(o_i)\leq 0\}.
\end{equation*}
The current policy then summarizes a compact empirical principle from both subsets:
\begin{equation*}
    p = \mathcal{P}_{\theta^{'}}
    \left(v,q,\mathcal{O}^{+},\mathcal{O}^{-}\right),
    \label{eq:pca_principle}
\end{equation*}
where $p$ captures a compact empirical principle by extracting reusable reasoning patterns from successful trajectories and contrasting them with common failure modes in incorrect trajectories (The detailed prompt for principle generation is provided in Appendix~\ref{app:Additional Implementation Details}).
To quantify how much this principle affects each reasoning step, we compare the original next-token distribution with the principle-conditioned distribution:
\begin{align*}
d_{i,t}^{\mathrm{pca}}
&=
D_{\mathrm{KL}}\Big(
\pi_{\theta'}(\cdot \mid v,q,o_{i,<t})
\,\Big\|\, \nonumber 
\pi_{\theta'}(\cdot \mid v,q,p,o_{i,<t})
\Big)
.
\label{eq:pca_kl}
\end{align*}
The KL divergence quantifies the degree of deviation between the original prediction and the fixed principle-conditioned guidance. 
A smaller $d_{i,t}^{\mathrm{pca}}$ indicates that the token prediction is better aligned with the principle-guided reasoning pattern.

\subsection{Shortcut-aware Contrastive Perturbation}
\label{ssec:scp}

Although PCA assigns finer reasoning-oriented credit, it does not determine whether the credited reasoning remains visually grounded during policy optimization.
In video RLVR, a policy may maintain confident and stable predictions after visual evidence is weakened, revealing a shortcut-driven tendency to rely on language priors rather than video content.
To expose such visual-invariant behavior, we introduce Shortcut-aware Contrastive Perturbation (SCP), which estimates response-level visual dependence by comparing policy distributions under original and perturbed video inputs.

We construct a counterfactual visual condition by applying a perturbation operator to the original video:
\begin{equation*}
    \tilde{v}=\mathcal{T}(v),
    \label{eq:scp_perturb}
\end{equation*}
where $\mathcal{T}(\cdot)$ weakens task-relevant visual evidence through operations such as frame replacement, region masking, or temporal corruption. 
Since the question $q$ and the generated prefix $o_{i,<t}$ are kept unchanged, this counterfactual condition isolates the effect of degraded visual evidence on the policy distribution.
Under this intervention, SCP measures the sensitivity of the policy distribution to weakened visual evidence. For each sampled response $o_i$, we define:
\begin{align}
d_{i,t}^{\mathrm{scp}}
&=
D_{\mathrm{KL}}\Big(
\pi_{\theta^{'}}(\cdot \mid v,q,o_{i,<t})
\,\Big\|\, \nonumber 
\pi_{\theta^{'}}(\cdot \mid \tilde{v},q,o_{i,<t})
\Big).
\label{eq:scp_kl}
\end{align}
The perturbed-video distribution is used as a fixed counterfactual reference, and the KL divergence measures the distributional discrepancy induced by weakening visual evidence. 
A larger $d_{i,t}^{\mathrm{scp}}$ indicates that the token prediction is more sensitive to visual perturbation, suggesting stronger dependence on evidence from the original video. 

\begin{table*}[htbp]
\centering
\begin{tabular}{lccccccc}
\toprule
\multirow{2}{*}{\textbf{Model}} & \multirow{2}{*}{\textbf{Frame}}
& \multicolumn{2}{c}{\textbf{Long Video}}
& \multicolumn{2}{c}{\textbf{General Video}}
& \multicolumn{2}{c}{\textbf{Complex Video}} \\
\cmidrule(lr){3-4} \cmidrule(lr){5-6} \cmidrule(lr){7-8}
& & \textbf{Video-MME} & \textbf{LVBench} & \textbf{TempCompass} & \textbf{MVBench} & \textbf{Video-MMMU} & \textbf{MMVU} \\
\midrule
\multicolumn{8}{c}{\textbf{\textit{Qwen3-VL-4B}}} \\
\hdashline
Base &64  &60.9  &38.7  &69.4  &60.9  &68.4  &63.8  \\
GRPO &64  &61.6  &40.2  &70.5  &61.8  &71.4  &66.1  \\
DAPO &64  &62.7  &40.9  &70.8  &\underline{63.5}  &71.2  &\underline{67.2}  \\
SDPO &64  &58.9  &33.0  &65.0  &58.6  &66.6  &65.0  \\
VISD &64  &\underline{63.3}  &\underline{41.3}  &\underline{72.3}  &62.8  &\underline{73.5}  &66.8 \\
\rowcolor{lightblue}
\textbf{BSPO (Ours)} 
&64  
&\textbf{66.7}$_{\textcolor{green!50!black}{+3.4}}$ 
&\textbf{44.3}$_{\textcolor{green!50!black}{+3.0}}$
&\textbf{78.0}$_{\textcolor{green!50!black}{+5.7}}$ 
&\textbf{68.5}$_{\textcolor{green!50!black}{+5.0}}$ 
&\textbf{77.5}$_{\textcolor{green!50!black}{+4.0}}$
&\textbf{73.0}$_{\textcolor{green!50!black}{+6.2}}$ 
\\
\midrule
\multicolumn{8}{c}{\textbf{\textit{Qwen3-VL-8B}}} \\
\hdashline
Base &64  &64.7  &40.2  &75.0  &65.3  &72.1  &69.7  \\
GRPO &64  &65.4  &40.8  &75.8  &66.0  &74.8  &70.2  \\
DAPO &64  &65.9  &41.9  &\underline{76.3}  &67.1  &\underline{76.2}  &\underline{72.0}  \\
SDPO &64  &62.5  &39.9  &70.8  &63.0  &68.0  &67.7  \\
VISD &64  &\underline{66.1}  &\underline{42.3}  &76.1  &\underline{67.9}  &75.9  &71.8  \\
\rowcolor{lightblue}
\textbf{BSPO (Ours)} &64  
&\textbf{68.2}$_{\textcolor{green!50!black}{+2.1}}$  
&\textbf{44.2}$_{\textcolor{green!50!black}{+1.9}}$ 
&\textbf{78.7}$_{\textcolor{green!50!black}{+2.6}}$  

&\textbf{69.9}$_{\textcolor{green!50!black}{+2.0}}$  
&\textbf{78.1}$_{\textcolor{green!50!black}{+1.9}}$    
&\textbf{73.3}$_{\textcolor{green!50!black}{+1.3}}$  \\
\bottomrule
\end{tabular}
\caption{Main results on video benchmarks. We compare BSPO with standard RLVR and OPSD baselines on two Qwen3-VL backbones across long-video understanding, general video understanding, and complex video reasoning tasks. All methods use the same number of frames. The best and second results are highlighted in bold and underline.}
\label{tab:main_result}
\end{table*}

\subsection{Bifocal Optimization}
\label{ssec:bifocal}

PCA and SCP provide two complementary distributional signals for regulating policy updates.
Since these signals are derived from distributional variations rather than outcome-level supervision, they should not override the preference specified by RLVR. 
Instead, they serve as auxiliary evidence to reallocate update strength across sampled responses, allowing bifocal optimization to preserve the sign of the trajectory-level advantage while adaptively modulating its magnitude according to reasoning sensitivity and visual dependence.

For each sampled response $o_i$, the detached PCA and SCP scores are normalized as:
\begin{equation*}
    \bar{d}_{i,t}^{m}
    =
    \frac{
    d_{i,t}^{m}-\mu_i^{m}
    }{
    \sigma_i^{m}
    },
    \qquad
    m\in\{\mathrm{pca},\mathrm{scp}\},
    \label{eq:bspo_token_norm}
\end{equation*}
where $\mu_i^{m}$ and $\sigma_i^{m}$ denote the mean and standard deviation of the corresponding score over all valid token positions in $o_i$, respectively.
This response-wise normalization avoids directly comparing token scores across rollouts with different lengths and generation dynamics. 
As a result, $\bar{d}_{i,t}^{m}$ reflects the relative importance of token $t$ within response $o_i$.
The normalized PCA and SCP scores are combined asymmetrically, since PCA favors small divergence from the principle-conditioned guidance whereas SCP favors large divergence from the perturbed-video reference. 
We therefore define the token-level reweighting factor as:
\begin{equation*}
    w_{i,t}^{\mathrm{bspo}}
    =
    1+\lambda
    \tanh
    \left(
    \frac{
    \operatorname{sgn}(A_i)
    \left(
    \bar{d}_{i,t}^{\mathrm{scp}}
    -
    \bar{d}_{i,t}^{\mathrm{pca}}
    \right)
    }{\tau}
    \right),
    \, 0<\lambda<1 ,
    \label{eq:bspo_weight}
\end{equation*}
where $\lambda$ controls the maximum reweighting range and $\tau$ controls the smoothness of the gate. Since $w_{i,t}^{\mathrm{bspo}}\in(1-\lambda,1+\lambda)$, the weight remains strictly positive and cannot reverse the update direction determined by $A_i$. 
For positive trajectories, this formulation increases the update weight for tokens that are closer to the empirical principle and more sensitive to visual perturbation. 
For negative trajectories, it increases the suppression of tokens that deviate from the empirical principle and remain insensitive to visual perturbations.
The final token-level advantage is obtained by multiplying the original trajectory-level advantage with the bifocal weight and clipping the result into a bounded interval for stability:
\begin{equation*}
    \widetilde{A}_{i,t}
    =
    A_i \cdot
    \operatorname{clip}\left(
    w_{i,t}^{\mathrm{bspo}},
    w_{\min},
    w_{\max}
    \right).
    \label{eq:bspo_advantage}
\end{equation*}
The BSPO objective follows the policy optimization form:
\begin{equation*}
\begin{split}
\mathcal{J}_{\text{BSPO}}(\theta) = \mathbb{E}_{\substack{(v,q)\sim\mathcal{D}\\\{o_i\}_{i=1}^G \sim \pi_{\theta'}(\cdot|v,q)}}
\Bigg[ \frac{1}{G}\sum_{i=1}^{G} \frac{1}{|o_i|}\sum_{t=1}^{|o_i|} \\
\min\!\left(w_{i,t}(\theta)\widetilde{A}_{i,t},\ \text{clip}\!\left(w_{i,t}(\theta), 1-\epsilon, 1+\epsilon\right)\widetilde{A}_{i,t}\right)\Bigg].
\end{split}
\end{equation*}

Through this design, BSPO preserves the verifier-anchored update direction of GRPO while reallocating credit at the token level. 
As a result, the policy optimization remains grounded in outcome rewards while becoming more selective about which tokens receive stronger updates.

\section{Experiment}

\begin{table*}[htbp]
\centering
\setlength{\tabcolsep}{7pt}
\renewcommand{\arraystretch}{1.12}
\begin{tabular}{lccccccc}
\toprule
\multirow{2}{*}{\textbf{Model}} 
& \multirow{2}{*}{\textbf{Frame}}
& \multicolumn{2}{c}{\textbf{Long Video}}
& \multicolumn{2}{c}{\textbf{General Video}}
& \multicolumn{2}{c}{\textbf{Complex Video}} \\
\cmidrule(lr){3-4} \cmidrule(lr){5-6} \cmidrule(lr){7-8}
& 
& \textbf{Video-MME} 
& \textbf{LVBench} 
& \textbf{TempCompass} 
& \textbf{MVBench} 
& \textbf{Video-MMMU} 
& \textbf{MMVU} \\
\midrule

BSPO 
& 64 & 68.2 & 44.2 & 78.7 & 69.9 & 78.1 & 73.3 \\

\midrule
\multicolumn{8}{l}{\textit{Ablation of BSPO components}} \\
\quad w/o PCA 
& 64 & 65.7 & 43.8 & 77.2 & 67.2 &76.0  & 70.2 \\
\quad w/o SCP 
& 64 & 65.9 & 43.5 & 76.1 & 66.5 & 77.2 & 71.0 \\

\midrule
\multicolumn{8}{l}{\textit{Ablation of SCP variants}} \\
\quad w/ Shuffle 
& 64 & 68.2 & 44.2 & 78.7 & 69.9 & 78.1 & 73.3 \\
\quad w/ Crop 
& 64 & 67.7 & 44.2 & 77.9 & 69.5 & 77.5 & 72.6 \\
\quad w/ Replace 
& 64 & 67.4 & 43.9 & 78.4 & 69.2 & 77.5 & 73.1 \\

\bottomrule
\end{tabular}
\caption{Ablation study. SCP and PCA are ablated from BSPO, and Shuffle, Crop, and Replace are different variants of SCP.}
\label{tab:ablation}
\end{table*}

\subsection{Experimental Setups}

\textbf{Backbone Models and Benchmarks.}
We conduct experiments on two representative video reasoning models, including Qwen3-VL-4B~\cite{qwen3vl} and Qwen3-VL-8B. 
To comprehensively evaluate the effectiveness of our method, we consider three categories of video understanding benchmarks. 
Video-MME~\cite{videomme} and LVBench~\cite{lvbench} assess long video understanding, emphasizing the aggregation of visual evidence across extended temporal contexts. 
TempCompass~\cite{temcompass} and MVBench~\cite{mvbench} evaluate general video understanding across diverse temporal, spatial, and semantic scenarios. 
MMVU~\cite{mmvu} and Video-MMMU~\cite{videommmu} target complex video reasoning, requiring multi-step inference grounded in fine-grained visual evidence.

\textbf{Baseline Methods.}
We compare BSPO with four representative baselines, including two standard RLVR methods and two on-policy self-distillation (OPSD) methods. 
For RLVR baselines, we include GRPO~\cite{deepseekr1}, which optimizes policies with group-normalized outcome rewards, and DAPO~\cite{dapo}, which improves large-scale RL training through decoupled clipping and dynamic sampling. 
For OPSD baselines, we compare with SDPO~\cite{sdpo} and VISD~\cite{visd}, where VISD is specifically designed for video reasoning, and both methods introduce self-generated dense supervision to alleviate the sparsity of outcome-level rewards.
For a fair comparison, all baselines are trained with the same backbone model, training data, reward function, and evaluation protocol.

\textbf{Implementation Details.}
We implement all methods with EasyR1~\cite{easyr1}. 
All experiments are conducted on 8 NVIDIA A100 GPUs with a global batch size of 128 with learning rate of 1e-6.
Models are trained for 1 epoch on 20K QA pairs sampled from Video-R1-260k~\cite{videor1}. 
For BSPO, we set $\lambda=0.5$, $\tau=1$, and clip the token-level reweighting factor to $[0.8,1.2]$ for all experiments. 
More details are provided in the Appendix~\ref{app:Additional Implementation Details}.

\subsection{Experiment Result}

Table~\ref{tab:main_result} reports the main results across six video understanding and reasoning benchmarks. 
Overall, BSPO achieves the best performance among all compared methods on both Qwen3-VL-4B and Qwen3-VL-8B, showing that the proposed optimization strategy generalizes across different model scales. 
Compared with standard RLVR methods such as GRPO and DAPO, BSPO brings more consistent improvements, suggesting that merely optimizing outcome-level rewards is insufficient for robust video reasoning. 
Compared with OPSD baselines such as SDPO and VISD, BSPO also shows stronger performance, indicating that its bifocal design better balances reasoning refinement and visual-dependence preservation.
Across all benchmark categories, BSPO achieves consistent improvements over existing baselines. 
The gains on long video and complex reasoning benchmarks suggest that preserving visual dependence is particularly beneficial when the model needs to aggregate temporally distributed evidence and perform fine-grained visual inference.

\subsection{Ablation Study}

Table~\ref{tab:ablation} presents the ablation results of BSPO. 
Removing PCA leads to a noticeable drop on reasoning-intensive benchmarks (e.g., MMVU), indicating that principle-guided credit assignment helps identify tokens that are truly relevant to the reasoning process. 
Removing SCP causes a larger degradation in visually grounded evaluation, especially under counterfactual settings (e.g., TempCompass). 
This suggests that reasoning-oriented credit assignment alone cannot prevent shortcut reliance when visual evidence is weakened. 
By penalizing predictions that remain insensitive to visual perturbations, SCP helps preserve visual dependence during RL training.

We further examine different SCP perturbation strategies, including shuffle, crop, and replacement. 
Compared with crop and replacement, shuffle provides a more balanced perturbation. 
Crop may remove task-relevant content too directly, while replacement can introduce additional distributional shifts. 
In contrast, shuffle preserves frame-level semantics but disrupts temporal order, thereby weakening temporal evidence without substantially altering the visual content.

\begin{figure*}[t]
    \centering
    \includegraphics[width= \linewidth]{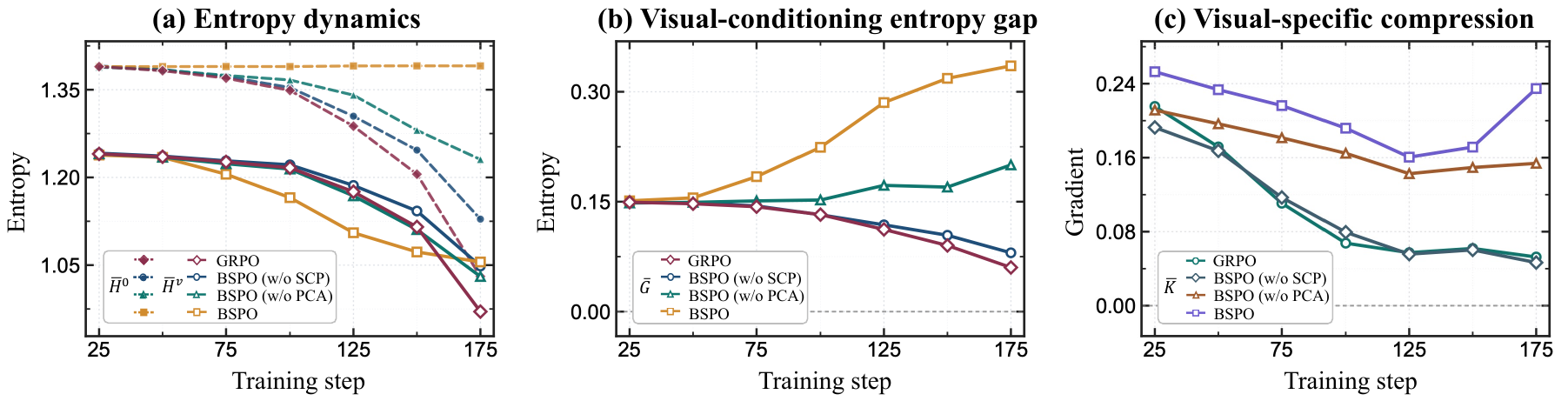}
    \caption{Perceptual uncertainty analysis of BSPO and its ablated variants. (a) Entropy dynamics compare the evolution of video-conditioned and text-only uncertainty. (b) The visual-conditioning entropy gap reflects the contribution of visual evidence to confidence formation. (c) Visual-specific compression measures the visual specificity of RL-induced uncertainty reduction. BSPO maintains stronger visual-evidence-dependent confidence formation than GRPO and its ablated variants.}
    \label{fig:experiment}
\end{figure*}

\subsection{Analysis}

\textbf{Perceptual Entropy Dynamics.}
Figure~\ref{fig:experiment}(a) and (b) presents the entropy dynamics during training. 
Consistent with our earlier observation, GRPO reduces the video-conditioned entropy $\bar{H}^{v}$ while the visual-conditioning entropy gap $\bar{G}$ gradually decreases, indicating that its confidence gain is weakly attributable to visual evidence. 
By contrast, BSPO produces a clearly different trend. 
Although $\bar{H}^{v}$ also decreases, the text-only entropy $\bar{H}^{0}$ remains relatively stable, leading to a steadily enlarged $\bar{G}$. 
This suggests that BSPO does not simply sharpen the output distribution, but makes the reduction in uncertainty more dependent on video conditioning. 
The ablated variants further support this interpretation. 
Removing SCP weakens the preservation of $\bar{G}$, showing that visual perturbation is crucial for preventing the model from bypassing video evidence. 
Removing PCA also degrades the trend, suggesting that reasoning-sensitive credit assignment helps convert visual dependence into effective policy updates.

\textbf{Visual-Specific Gradient Compression.}
Figure~\ref{fig:experiment}(c) further compares visual-specific compression $\bar{K}$, which measures whether RL-induced sharpening is more aligned with video-conditioned entropy reduction than text-only compression. 
GRPO shows a rapid decline in $\bar{K}$, confirming that its update direction becomes increasingly dominated by text-prior compression. 
In contrast, BSPO maintains a substantially higher $\bar{K}$ throughout training and recovers at later stages, indicating that its updates remain more visually specific. 
This improvement mainly comes from the two components.
Without SCP, $\bar{K}$ drops close to the GRPO trajectory, suggesting that perturbation-based visual dependence is the main signal for suppressing shortcut-invariant updates. 
Without PCA, $\bar{K}$ remains higher than that of GRPO but lower than that of the full BSPO, suggesting that reasoning-aware credit assignment further strengthens the visual specificity of policy updates.
Together, these results show that BSPO changes confidence formation by aligning RL updates with visually grounded uncertainty reduction.

\begin{figure}[t]
    \centering
    \includegraphics[width=\linewidth]{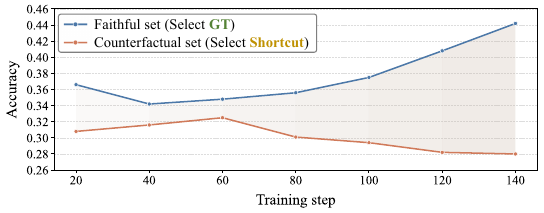}
    \caption{Shortcut robustness performance. Accuracy trends on the faithful and counterfactual sets during training.}
    \label{fig:robustness}
\end{figure}

\textbf{Shortcut Robustness under Counterfactual Videos.}
We further evaluate robustness under counterfactual videos, where the visual evidence no longer supports the text-prior answer. 
In contrast to the narrowing separation observed under GRPO in Figure~\ref{fig:motivation}, Figure~\ref{fig:robustness} shows that BSPO progressively enlarges the performance gap between the faithful and counterfactual subsets during training. 
Specifically, BSPO improves accuracy on visually valid examples while suppressing shortcut answers once their supporting evidence is invalidated. 
This reversed trend indicates that BSPO mitigates the shortcut amplification identified and strengthens dependence on visual evidence rather than reinforcing textual priors.
\section{Related Work}
\textbf{RLVR for Video Reasoning.}
RLVR has recently been adopted for multimodal and video reasoning by using rule-based outcome rewards to scale post-training without dense human annotations~\cite{deepseekr1,dapo}. 
Existing works mainly improve video reasoning through GRPO-style optimization, video-grounded data construction, or task-specific rewards for temporal and spatio-temporal reasoning~\cite{videor1,rewatchr1,videostr,longvilar1,glm41v,onethinker,keye1.5,videochatr15}. 
However, these methods largely emphasize performance improvement and provide limited analysis of whether the resulting gains are genuinely grounded in visual evidence.
In contrast, our work focuses on a failure mode of video RLVR where outcome-level rewards may improve accuracy while amplifying shortcut predictions that are not supported by visual evidence. 
We analyze this from a perceptual uncertainty perspective and design BSPO to preserve visual dependence during RL training.

\textbf{OSPD for Video Reasoning.}
OSPD introduces self-generated signals from the model's own rollouts to provide denser supervision than final-answer rewards~\cite{sdpo,rlsd,rlcsd}. 
Recent OSPD methods improve reasoning or visual perception by exploiting policy differences, privileged guidance, or self-distilled token-level feedback~\cite{videoopd,visd}. 
These approaches alleviate the coarse credit assignment induced by outcome-level rewards and provide more informative supervision for intermediate reasoning steps. 
However, most of them primarily focus on improving optimization efficiency or reasoning quality, without explicitly examining whether the resulting confidence remains grounded in visual evidence. 
BSPO differs by coupling self-distillation with counterfactual visual perturbation, so that policy updates are jointly guided by reasoning sensitivity and visual dependence.
\section{Conclusion}
In this paper, we reveal shortcut amplification in video RLVR and analyze its training dynamics through perceptual uncertainty. 
Our entropy and gradient analyses show that outcome-level RL can make models more confident while making uncertainty reduction less attributable to visual evidence. 
This exposes a gap between final-answer accuracy and genuine visual grounding. 
To mitigate this issue, we propose BSPO, which combines principle-guided credit assignment with shortcut-aware contrastive perturbation to promote reasoning-sensitive and visually dependent updates. 
Extensive experiments show that BSPO improves performance and strengthens visual faithfulness during RL training.

\bibliography{main}

\clearpage
\appendix

\section{Additional Implementation Details}
\label{app:Additional Implementation Details}
\begin{figure*}[th]
    \centering
    \includegraphics[width=\linewidth]{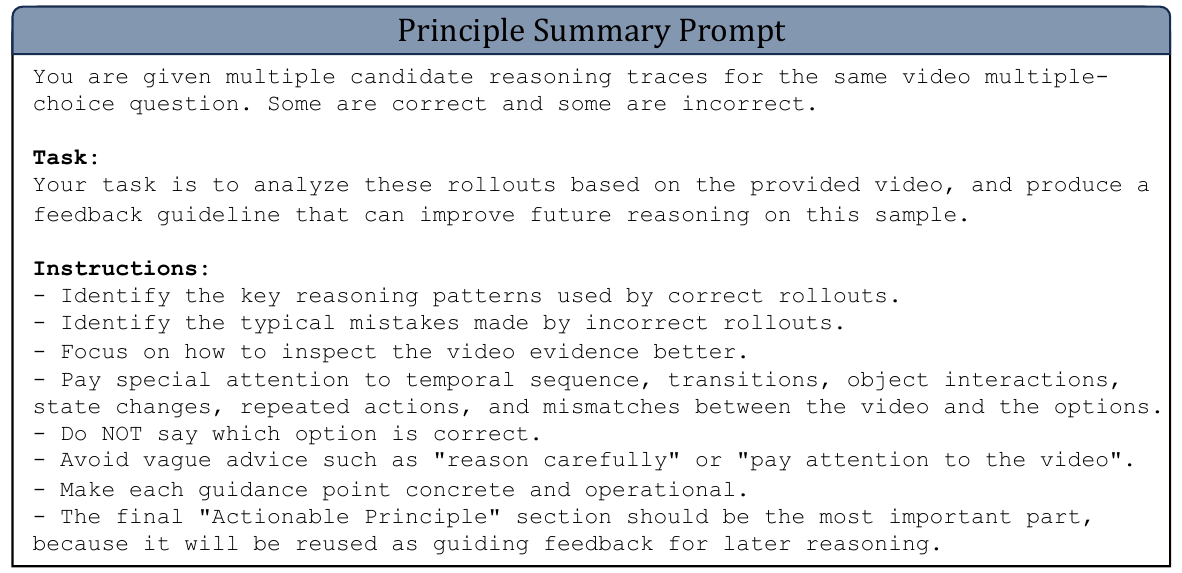}
    \caption{Prompt used to summarize empirical reasoning principles from correct and incorrect rollouts for principle-guided credit assignment (PCA).}
    \label{fig:prompt}
\end{figure*}

For SCP, we randomly select $20\%$ of the input frames and apply one of three perturbations. Shuffle randomly permutes the selected frames, crop retains a randomly cropped region covering $50\%$ of each selected frame and resizes it to the original resolution, and replacement replaces the selected frames with blank frames. 
Shuffle is used as the default perturbation unless otherwise specified. 
The entropy and gradient analyses are conducted on Video-MME using Qwen3-VL-8B, where all checkpoints are evaluated on the same samples and ground-truth responses under teacher forcing to ensure a controlled comparison.
The prompt used for principle summary is shown in Figure~\ref{fig:prompt}.
The reported results are averaged over 3 independent runs with different seeds.

\section{Further Theoretical Analysis}

\subsection{Interpretation of the Visual-conditioning Entropy Gap}

We further clarify the interpretation of the visual-conditioning entropy gap. 
The quantity $\bar{G}$ does not directly measure the amount of visual information encoded in the hidden representation. Instead, it measures the extent to which conditioning on the video reduces uncertainty in the model's next-token predictive distribution under an otherwise identical textual context. Therefore, $\bar{G}$ should be interpreted as an output-level measure of visual contribution to confidence formation.

Under the controlled teacher-forcing setting, the question and response prefix are identical for the video-conditioned and text-only distributions. Hence, the difference between $\bar{H}^{v}$ and $\bar{H}^{0}$ is induced only by the presence of the video condition. A decreasing $\bar{G}$ indicates that the additional uncertainty reduction provided by the video becomes weaker relative to the text-only context. When this occurs together with a decreasing video-conditioned entropy, $\bar{H}^{v} \downarrow$ and $\bar{G} \downarrow$, the model becomes more confident, while a smaller proportion of this confidence can be attributed to visual conditioning.

Accordingly, perceptual uncertainty collapse in this work does not refer to a directly observed collapse of an explicit visual latent variable. It denotes an observable output-level failure mode in which predictive uncertainty is increasingly compressed without a corresponding increase in the uncertainty reduction attributable to visual evidence.

\subsection{Teacher-forced Gradient Analysis as a Controlled Update Probe}

The actual policy update in RLVR is computed on sampled trajectories, whereas our gradient analysis is evaluated on a fixed ground-truth trajectory. We clarify that the latter is used as a controlled probe of the local effect of a policy-gradient direction, rather than as an exact reconstruction of every stochastic training update.

For a generic token $y_t$ with context $x_t$, consider the local log-likelihood objective:
\[
J_t(\theta)
=
A_t
\log p_{\theta}(y_t \mid x_t),
\]
which induces the first-order update:
\[
\Delta\theta_t
=
\eta A_t
\nabla_{\theta}
\log p_{\theta}(y_t \mid x_t).
\]
For any differentiable uncertainty functional $H_t(\theta)$, its first-order change under this update is:
\[
\Delta H_t
=
H_t(\theta+\Delta\theta_t)-H_t(\theta)
\approx
\nabla_{\theta}H_t(\theta)^{\top}\Delta\theta_t.
\]
Substituting the policy-gradient update gives:
\[
\Delta H_t
\approx
\eta A_t
\nabla_{\theta}H_t(\theta)^{\top}
\nabla_{\theta}
\log p_{\theta}(y_t \mid x_t).
\]

This relation holds for any fixed token and prefix, independently of whether they are obtained from a sampled rollout or a ground-truth trajectory. The role of teacher forcing is therefore to eliminate trajectory-level variation when comparing checkpoints and input conditions. In particular, using the same ground-truth token $y_t^{\mathrm{gt}}$ and the same prefix $y_{<t}^{\mathrm{gt}}$ ensures that changes in the measured gradient geometry arise from changes in the policy parameters or conditioning modality, rather than from
different sampled reasoning paths. Thus, the teacher-forced quantity $-A_t
\nabla_{\theta}H_t(\theta)^{\top}
\nabla_{\theta}
\log p_{\theta}
\left(
y_t^{\mathrm{gt}}
\mid x_t
\right)$
should be understood as a local update-response probe. It measures whether increasing the likelihood of the ground-truth continuation would locally compress or expand predictive uncertainty at the current checkpoint. This controlled formulation is particularly suitable for comparing video-conditioned and text-only uncertainty-reducing directions under identical semantic supervision.

\end{document}